\documentclass{article} 
\usepackage{iclr2026_conference,times}

\usepackage{amsmath,amsfonts,bm}

\def\eqref#1{equation~\ref{#1}}

\def\1{\bm{1}}

\DeclareMathAlphabet{\mathsfit}{\encodingdefault}{\sfdefault}{m}{sl}
\SetMathAlphabet{\mathsfit}{bold}{\encodingdefault}{\sfdefault}{bx}{n}

\definecolor{citecolor}{HTML}{0071bc}
\usepackage[pagebackref=false,breaklinks=true,letterpaper=true,colorlinks,citecolor=citecolor,bookmarks=false]{hyperref}
\usepackage{url}            

\usepackage{graphicx}
\usepackage{enumitem}
\usepackage{booktabs,caption,tabularx,array,multirow}
\usepackage{placeins}
\usepackage{makecell}
\usepackage{amsthm}
\usepackage{multirow}
\usepackage{booktabs}
\usepackage{wrapfig}
\usepackage{caption}
\usepackage{amssymb}
\usepackage{diagbox}
\usepackage{bbding}
\usepackage{subcaption}
\usepackage{ulem}
\usepackage{pifont}
\usepackage{bbm}
\usepackage{tcolorbox}
\usepackage{lipsum}
\usepackage{graphicx}
\usepackage[table,x11names]{xcolor}
\usepackage{changepage}
\usepackage{tikz}
\usepackage{longtable}

\definecolor{darkgreen}{rgb}{0.0, 0.8, 0.0}
\definecolor{acolor}{RGB}{84, 130, 53}
\definecolor{bcolor}{RGB}{84, 130, 53}
\definecolor{ccolor}{RGB}{84, 130, 53}
\definecolor{dcolor}{RGB}{84, 130, 53}

\title{NANO3D: A Training-Free Approach for Efficient 3D Editing Without Masks}

\author{
  \textbf{Junliang Ye}$^{1}$\footnotemark[1]\quad\quad
  \textbf{Shenghao Xie}$^{2,1}$\thanks{Equal contribution}\quad\quad
  \textbf{Ruowen Zhao}$^{1}$\quad\quad
  \textbf{Zhengyi Wang}$^{1}$\quad\quad
  \textbf{Hongyu Yan}$^{4}$\\
  \textbf{Wenqiang Zu}$^{5,2}$\quad\quad
  \textbf{Lei Ma}$^{2}$\footnotemark[2]\quad\quad\quad\quad
  \textbf{Jun Zhu}$^{1,3}$\thanks{Corresponding author.} \quad\quad \\
  $^{1}$Tsinghua University\quad $^{2}$Peking University\quad $^{3}$ShengShu\quad $^{4}$HKUST\quad$^{5}$CASIA
}

\newcommand{\ie}{{\textit{i.e.}}}

\newcommand{\eg}{{\textit{e.g.}}}

\iclrfinalcopy 
\begin{document}

\newtheorem{theorem}{Theorem}[section]
\newtheorem{proposition}[theorem]{Proposition}

\maketitle

\vspace{-10mm}
\begin{figure}[h]
\begin{center}
\includegraphics[width=\linewidth]{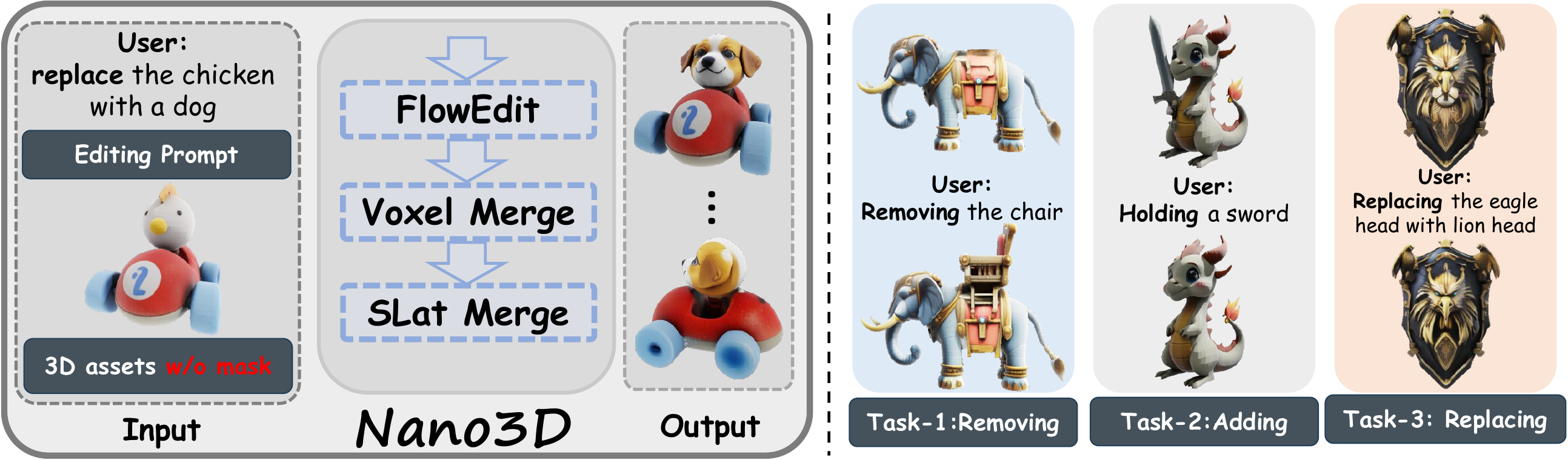}
\end{center}
\caption{\textbf{Highly-consistent 3D objects edited by Nano3D.} Our framework supports a range of training-free and part-level tasks especially on shape, including removal, addition, and replacement, while only requiring users to provide source 3D objects and instructions, without any mask.}
\label{fig:head}
\end{figure}

\begin{abstract}
3D object editing is essential for interactive content creation in gaming, animation, and robotics, yet current approaches remain inefficient, inconsistent, and often fail to preserve unedited regions. Most methods rely on editing multi-view renderings followed by reconstruction, which introduces artifacts and limits practicality. To address these challenges, we propose \textbf{Nano3D}, a training-free framework for precise and coherent 3D object editing without masks. Nano3D integrates FlowEdit into TRELLIS to perform localized edits guided by front-view renderings, and further introduces region-aware merging strategies, Voxel/Slat-Merge, which adaptively preserve structural fidelity by ensuring consistency between edited and unedited areas. Experiments demonstrate that Nano3D achieves superior 3D consistency and visual quality compared with existing methods. Based on this framework, we construct the first large-scale 3D editing datasets \textbf{Nano3D-Edit-100k}, which contains over 100,000 high-quality 3D editing pairs. This work addresses long-standing challenges in both algorithm design and data availability, significantly improving the generality and reliability of 3D editing, and laying the groundwork for the development of feed-forward 3D editing models. \\
\textbf{Project Page}:\url{https://jamesyjl.github.io/Nano3D}
\end{abstract}
\section{Introduction}~\label{sec:introduction}

Generative models for 3D asset creation have made tremendous progress~\cite{lai2025hunyuan3d2.5, chen20253dtopiaxl, chen2024meshanythingv2, wang2023prolificdreamer}, leading to widespread applications across entertainment, robotics, and healthcare. In particular, recent rectified flows (reflows)~\cite{liu2022rectifiedflow}, such as TRELLIS~\cite{xiang2025trellis}, achieve high-quality 3D object generation by embedding heterogeneous representations into a unified latent space while explicitly disentangling geometry and appearance. Beyond generation, editing (\ie, revising the intended region while keeping other regions unchanged) is also valuable as users usually need to refine existing assets rather than regenerate entirely new ones, which requires multiple unpredictable iterations to obtain a satisfactory result. For example, TRELLIS can generates diverse plausible appearances easily with style-modified text or image prompts, such as texture and material, but fail to reliably repeat identical geometries.

In image editing, an increasing number of powerful models have recently emerged, including GPT-4o~\cite{hurst2024gpt4o}, Flux.1 Kontext~\cite{labs2025flux1kontext}, and Nano Banana~\cite{Fortin2025Gemini2.5Flashimage}. A closer look at the evolution of these models reveals a clear three-stage development paradigm. Stage 1 introduced training-free image editing algorithms~\cite{hertz2022prompt2prompt}, which demonstrated the feasibility of editing without model finetuning. Stage 2 focused on the automatic construction of large-scale, high-quality paired editing datasets, providing the foundation for supervised learning~\cite{brooks2023instructpix2pix}. Stage 3 leveraged these datasets to train feedforward image editing models capable of real-time inference and high fidelity generation.

In contrast, 3D object editing still remains bottlenecked in the initial stage (\ie, algorithm). Specifically, existing methods, such as those based on Score Distillation Sampling (SDS)~\cite{sella2023voxe} or the ``multi-view editing then reconstruction'' paradigm~\cite{qi2024tailor3d}, struggle to maintain consistency across views or attributes and usually demand time-consuming optimization. This leaves us wondering: \textit{can 3D objects be edited versatilely, efficiently and consistently in a training-free manner using only pretrained generative models, as achieved in 2D images?} Resolving this problem will allow 3D object editing to enter a virtuous cycle of data expansion and training models capable of flexible asset customization, thereby accelerating the whole field toward maturity like 2D images.

We propose Nano3D, a training-free 3D editing algorithm designed for constructing paired 3D editing datasets. Drawing inspiration from the training-free 2D editing method FlowEdit~\cite{kulikov2024flowedit}, Nano3D leverages the first stage of TRELLIS to generate an iterative trajectory from input to edited voxel representations, thereby enabling efficient training-free 3D editing.

To further enhance source consistency between the original and edited objects, we introduce a region-aware merging strategy, Voxel/Slat-Merge, applied after TRELLIS's two-stage geometry and appearance editing. Based on simple connectivity analysis, this strategy adaptively identifies modified voxel regions in the edited 3D object and integrates them back into the original object. This effectively merges the edited content while preserving the structure of unedited regions.

Building on the Nano3D algorithm, we design an efficient pipeline for large-scale construction of 3D editing datasets and generate a high-quality dataset of 100,000 samples——\textbf{Nano3D-Edit-100k}. Our work addresses two long-standing gaps in the 3D editing domain—the lack of training-free editing algorithms and the absence of large-scale datasets—thereby laying a solid foundation for the third stage of 3D editing: training feedforward models under 3D editing supervision.

Overall, our contributions can be summarized as follows:
\begin{itemize}[leftmargin=*]
\item We make the first attempt to introduce FlowEdit to 3D object editing, demonstrating that the powerful priors of 3D object generative models can also support effective training-free editing (like 2D images)
\item We propose Voxel/Slat-Merge, a region-aware merging strategy that automatically preserves source consistency in the non-edited regions of 3D objects.
\item We develop a user-friendly 3D editing framework, \textbf{Nano3D}, which achieves state-of-the-art editing performance without requiring any manual masks.
\item Building upon Nano3D, we curate the first large-scale 3D editing dataset \textbf{Nano3D-Edit-100k}, comprising over 100,000 high-quality samples to support further research and development.
\end{itemize}

\section{Related Work}~\label{sec:related_work}
\vspace{-8mm}
\subsection{2D Image Editing}~\label{sec:2dimageediting}
With the advent of large-scale 2D generative models, image editing has shifted from manual pixel-level operations to controllable semantic-level manipulation. Early approaches modify noisy latents via inversion to balance new details with original structures~\cite{meng2021sdedit, mokady2023nulltextinversion, abdal2019image2stylegan}, while others finetune generative models on curated editing pairs to enable instruction following~\cite{brooks2023instructpix2pix, wei2024omniedit, sheynin2024emuedit}. Localized editing has also been explored through attention map manipulation~\cite{hertz2022prompt2prompt, tumanyan2023plugandplay, couairon2022diffedit}, and adapters have been introduced to inject additional conditions for enhanced controllability~\cite{ye2023ipadapter, ju2024brushnet, mou2024t2iadapter}. More recently, rectified flows (reflow)~\cite{liu2022rectifiedflow, esser2024sd3} have enabled high-fidelity synthesis with few sampling steps. To support reflow-based editing, RFSolver~\cite{wang2024rfsolver} approximates ODEs via higher-order Taylor expansion while preserving structures through attention replacement, whereas FlowEdit~\cite{kulikov2024flowedit} introduces an inversion-free strategy by interpolating between sampled noise and the source image.
\subsection{Related Work about 3D Object Generation}~\label{sec:3dobjectgeneration}
Over the past years, 3D object generation has been pursued under several paradigms.Generative Adversarial Net (GAN)-based approaches are among the earliest to directly model 3D distributions~\cite{goodfellow2014gan, chan2022efficientgeometryaware3dgenerative, gao2022get3dgenerativemodelhigh,weng2024scalingmeshgenerationcompressive, zheng2022sdfstyleganimplicitsdfbasedstylegan}. Diffusion models later become popular by treating 3D representations as denoising tasks, which provides improved training stability and superior generative fidelity~\cite{chen2023fantasia3d,wang2023prolificdreamer,ye2024dreamrewardtextto3dgenerationhuman, 11164374,wang2024animatabledreamertextguidednonrigid3d, liu2025reconxreconstructscenesparse}. At the same time, autoregressive (AR) models~\cite{weng2024pivotmesh,zhao2025deepmeshautoregressiveartistmeshcreation,chen2024meshanythingv2, wei2025octgptoctreebasedmultiscaleautoregressive, chen2025autopartgenautogressive3dgeneration} enable fine-grained conditional control and structured synthesis through ordered generation. More recently, reflow models formulate 3D generation as a continuous ODE-based linear transformation, enabling faster inference with competitive performance~\cite{xiang2025trellis, hunyuan3d2025hunyuan3d2.1, li2025triposg,chen2025ultra3d,ye2025hi3dgen}.
\subsection{3D Object Editing}~\label{3dobjectediting}
Compared to 2D image editing, maintaining spatial consistency is substantially more challenging in 3D. Many approaches adopt score distillation sampling (SDS)~\cite{poole2022dreamfusion} to optimize 3D representations using gradients from pretrained 2D diffusion models~\cite{sella2023voxe, li2024focaldreamer, chen2024shapeditor, palandra2024gsedit, chen2023plasticine3d}. Others edit multi-view images and reconstruct them with large reconstruction models (LRMs)~\cite{qi2024tailor3d, chen2024mvedit, barda2025instant3dit, huang2025edit360, erkocc2025preditor3d, bar2025editp23, zheng2025pro3deditor, li2025cmd, gao20243d}, or directly manipulate triplanes as a bridge between 2D and 3D~\cite{kathare2025instructive3d, bilecen2025reference}. Inspired by InstructPix2Pix~\cite{brooks2023instructpix2pix}, several works construct paired 3D editing datasets for supervised training~\cite{ye2025shapellmomni, xu2023instructp2p}. To enable finer control, diverse conditions such as sketches~\cite{mikaeili2023sked, liu2024sketchdream, guillard2021sketch2mesh}, part-level masks~\cite{chen2025partgen, yang2025holopart, yang2025omnipart,li2025voxhammer}, and point-based dragging~\cite{chen2024mvdrag3d, xie2023dragd3d, lu2025dreamart} have been explored. More recently, rectified flows (reflow)~\cite{zhao2025hunyuan3d2.0, li2025triposg, he2025sparseflex} achieved large-scale 3D generation and zero-shot appearance editing, yet still face bottlenecks in shape modification. In this work, we unlock their potential for versatile and consistent 3D editing in a training-free and user-friendly manner.
\section{Preliminary}~\label{sec:preliminary}
\vspace{-8mm}
\subsection{FlowEdit}
FlowEdit is a text-guided image editing method tailored for text-to-image flow models. It is characterized by being inversion-free, optimization-free, and model-agnostic. Rather than relying on traditional inversion-reconstruction processes that often introduce distortion, FlowEdit constructs an ordinary differential equation (ODE) trajectory in the latent space from the source prompt to the target prompt. This trajectory enables direct evolution of image representations over the velocity field. By leveraging a weighted combination of the source and target velocity fields, FlowEdit ensures a shorter editing path and stronger structural preservation throughout the editing process.

\subsection{Trellis}
Trellis represents a 3D asset using Structured Latents (SLAT) — a set of activated voxels, each associated with a local latent vector, denoted as ${(z_i, p_i)}$. Here, each $z_i \in \mathbb{R}^C$ jointly encodes both geometric and appearance information in a compact yet expressive manner. The inference pipeline consists of two successive stages. In the Structure Prediction (ST) stage, a $64^3$ voxel grid is employed to estimate occupancy, thereby producing a sparse structural representation of the asset. In the Sparse Latent (SLAT) stage, local latent vectors are further inferred for the voxels identified in the first stage, capturing richer semantic and geometric details. Finally, a powerful SLAT-VAE decoder reconstructs the complete 3D asset from the predicted latent representations with high fidelity.
\section{Method}~\label{sec:method}
\vspace{-5mm}
\begin{figure}[ht]
\begin{center}
\includegraphics[width=\linewidth]{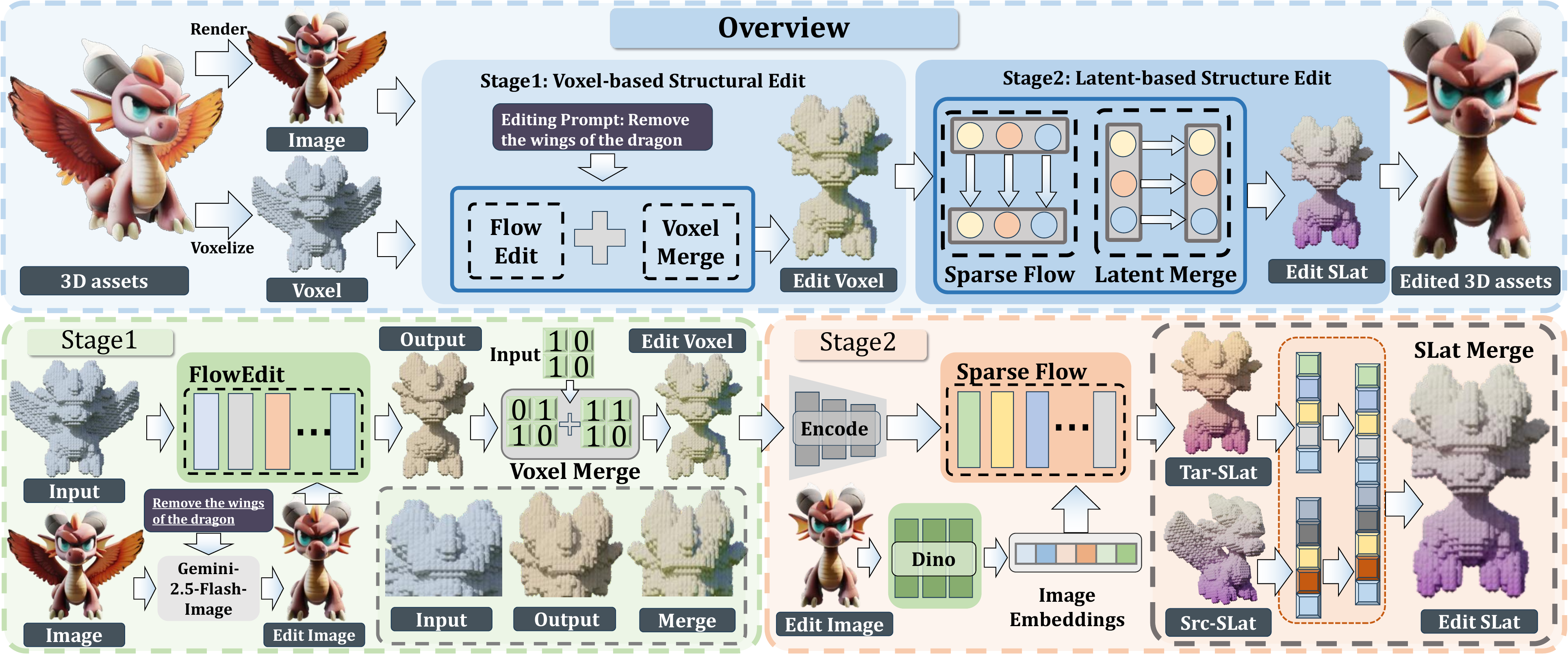}
\end{center}
\caption{\textbf{The Nano3D pipeline.} The original 3D object is voxelized and encoded into sparse structure and structured latent respectively. Stage 1 modifies geometry via Flow Transformer with FlowEdit, guided by Nano Banana–edited images. Stage 2 generates structured latents with Sparse Flow Transformer, supporting TRELLIS-inherent appearance editing. Voxel/Slat-Merge further ensures consistency across both stages before decoding the final 3D object.}
\label{fig:pipeline}
\end{figure}

\vspace{-6mm}

\subsection{Overview}
A common approach is to edit rendered images of a 3D object and reconstruct it with a generative model, but this often breaks geometric consistency. To address this, we introduce FlowEdit into the first-stage generation of TRELLIS (Sec.~\ref{sec:flowedits}). To further ensure geometric and appearance consistency, we propose Voxel/Slat-Merge (Sec.~\ref{sec:merge}), which detects edited regions and integrates them with unedited ones. Finally, we present a training-free, user-friendly pipeline (Sec.~\ref{sec:pipeline}) that also supports large-scale dataset construction. By combining FlowEdit with Voxel/Slat-Merge, Nano3D achieves geometrically consistent and semantically faithful 3D object editing within TRELLIS.

\subsection{FlowEdit}~\label{sec:flowedits}
Inspired by FlowEdit’s success in 2D image editing, we extend it to 3D object editing by integrating it into TRELLIS stage 1, leveraging the pretrained generative prior to establish an editing path between source and target objects instead of starting from noise. Specifically, we use input 3D object’s front view as source condition and Nana-Banana–edited image as the target. From the voxelized input, FlowEdit’s iterative inference generate the edited voxel output.

\subsection{Voxel/Slat-Merge}~\label{sec:merge}
\textbf{Voxel-Merge} We observe that voxels produced by FlowEdit do not always preserve structural consistency between the input and the edited asset. For example, when editing a dragon to remove its wings, the resulting voxels may not only modify the wings but also inadvertently alter other unrelated regions. To effectively mitigate this issue, we introduce a region-aware merging strategy, Voxel-Merge, applied at the end of the geometry editing stage. This method explicitly defines a difference map $g$ via a voxel-wise XOR operation between sparse structures:
\begin{equation}
g(i) = s_{\text{src}}(i) \oplus s_{\text{tgt}}(i) =
\left\{
\begin{array}{ll}
1, & \text{if } s_{src}(i) \neq s_{tgt}(i), \\[6pt]
0, & \text{if } s_{src}(i) = s_{tgt}(i).
\end{array}
\right.
\quad \forall i
\end{equation}
Under different connectivities (\eg, 6/18/26-neighborhoods), $g$ is decomposed to several components $\{ g_j \}_{j=1}^C$, where $\lvert g_j \rvert$ is the number of voxels. After the descending sorting, we adaptively select top-$k$ components and construct the flip mask $m = \{ m_j \}_{j=1}^C$, where $m_j = \{1\}^{\lvert g_j \rvert}$ if $j \in \mathcal{I}_k$ and $m_j = \{0\}^{\lvert g_j \rvert}$ otherwise. Alternatively, components larger than a threshold $\tau$ can be selected, \ie, $m_j = \{1\}^{\lvert g_j \rvert}$ if $\lvert g_j \rvert > \tau$, which ignores small noisy and irrelevant regions, ensuring only significant discrepancies (\ie, edited regions) are masked. Then we perform $s_{src} \oplus M$ to transfer edited regions of $s_{tgt}$ onto the source 3D object, and inherently ensure non-edited regions unchanged. 

\textbf{SLat-Merge} After applying Voxel-Merge, we obtain voxels that are fully consistent before and after editing. We then feed the edited image together with the merged voxels into the second stage of TRELLIS to generate SLat representations. To ensure that the generated SLat features remain consistent with those encoded from the original 3D asset, we further introduce SLat-Merge. Specifically, this module continues to utilize the previously defined mask to perform $z_{src} \oplus M$, thereby merging the edited and non-edited regions within the structured latent space.

\subsection{Nano3D}~\label{sec:pipeline}
As illustrated in Fig.~\ref{fig:pipeline}, Nano3D builds upon TRELLIS to enable decoupled geometry and appearance editing of 3D objects. The input object is voxelized and, along with DINOv2~\cite{oquab2023dinov2} features, encoded into a structured latent representation via a VAE~\cite{kingma2013vae}. Meanwhile, we use Nano Banana with the front view of a 3D asset and editing instructions as input to generate the edited front view. In TRELLIS-Stage 1, we replace the standard flow iteration with FlowEdit, which takes the source object’s voxel and the before/after front views as input, and outputs the edited voxel. We then apply Voxel-Merge to ensure geometric consistency. In TRELLIS-Stage 2, the edited voxel and edited front view jointly guide TRELLIS to generate the final SLat. At this stage, we further adopt Slat-Merge to guarantee both geometric and texture consistency. Finally, the edited SLat is decoded by the VAE to reconstruct the target 3D object.

\textbf{Data Construction Pipeline.} As illustrated in Fig.~\ref{fig:pipeline-data}, we extend \textbf{Nano3D} by constructing a complete and streamlined \textbf{3D editing data generation pipeline}. The process consists of the following stages:
\begin{figure}[h]
\begin{center}
\includegraphics[width=\linewidth]{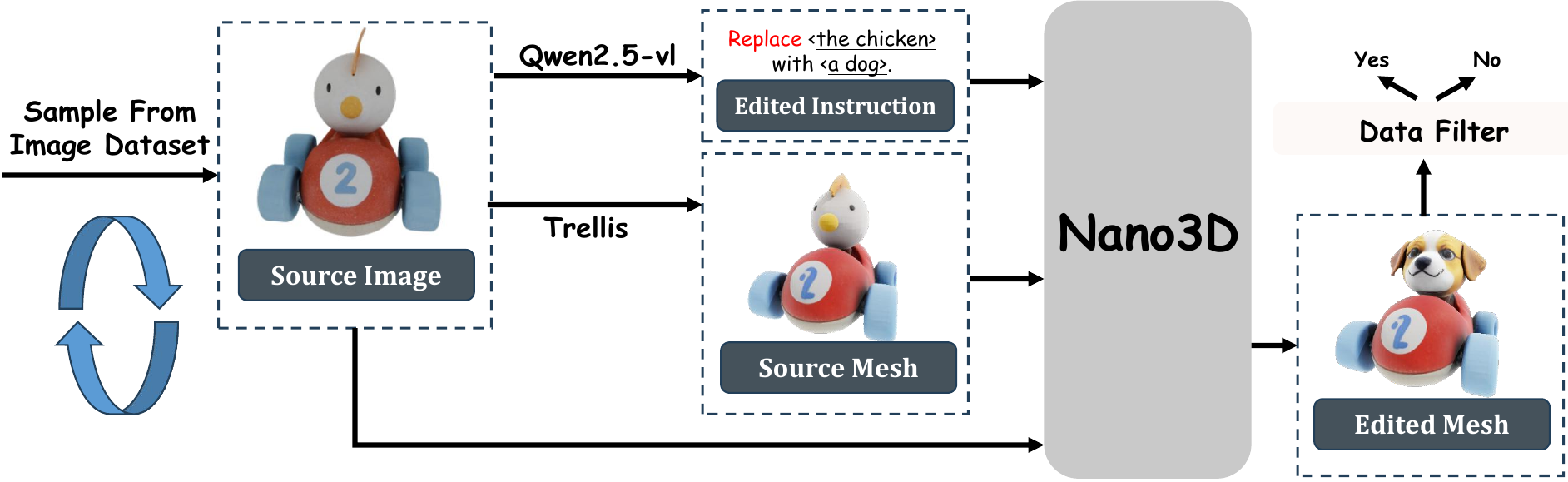}
\end{center}
\caption{\textbf{Data Construction Pipeline.} The figure shows our pipeline. We first sample images from the dataset and prompt Qwen2.5-VL to generate editing instructions by completing templates. Trellis then generates 3D meshes from the images. Finally, the image, instruction, and mesh are fed into Nano3D, and the resulting 3D assets are filtered for quality.}
\label{fig:pipeline-data}
\end{figure}
\begin{enumerate}
\item \textbf{Image Sampling from Existing Datasets}: We sample views from publicly available 3D asset datasets~\cite{xiang2025trellis, deitke2022objaverseuniverseannotated3d}. For each asset, the frontal view is selected as the editing target.
    
    \item \textbf{Instruction Generation via VLM}: An editing instruction is automatically generated using the vision-language model \textbf{Qwen-VL-2.5}~\cite{bai2025qwen25vltechnicalreport}, based on three predefined prompt templates:
    \begin{itemize}
        \item \textbf{Add}: \texttt{add <something> to <somewhere>}
        \item \textbf{Remove}: \texttt{remove <something> in <somewhere>}
        \item \textbf{Replace}: \texttt{replace <something> with <something>}
    \end{itemize}
    The model fills in these templates with visual context from the image to produce diverse and semantically grounded instructions.
    
    \item \textbf{3D Asset Generation via TRELLIS}: Given the selected image, we use \textbf{TRELLIS} to reconstruct the corresponding 3D asset. Instead of using the original mesh, we choose to regenerate the source mesh via TRELLIS for two reasons: (1) obtaining the structured latent (sLat) from the original mesh requires rendering $\sim$150 views, which is inefficient; (2) the reconstructed sLat still diverges from the original mesh due to the inherent loss in TRELLIS's VAE encoding. Using the TRELLIS-reconstructed mesh ensures consistency and reduces computational overhead.

    \item \textbf{Image Editing via Nano-Banana or Flux-Kontext}: The generated instruction is input into \textbf{Nano-Banana} or \textbf{Flux-Kontext} to synthesize the edited target image.
    
    \item \textbf{3D Editing via Nano3D}: The original 3D asset, the source image, and the edited image are fed into \textbf{Nano3D}, which outputs an edited 3D asset.
\end{enumerate}
\section{Evaluation}~\label{sec:evaluation}
\vspace{-10mm}
\begin{figure}[h]
\begin{center}
\includegraphics[width=\linewidth]{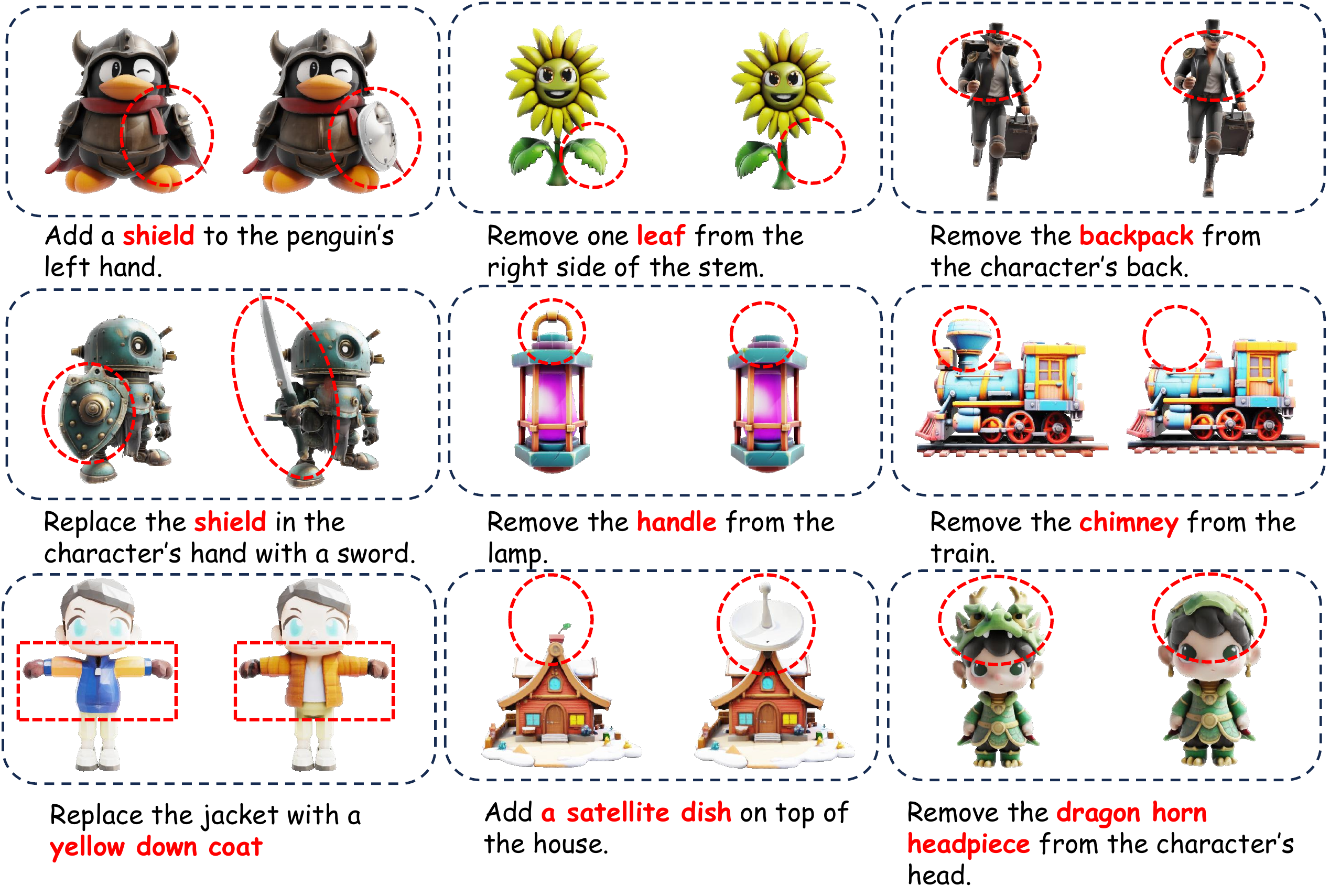}
\end{center}
\caption{\textbf{Qualitative results.} We present three edit types—object removal, addition, and replacement. In each case, Nano3D confines changes to the target region (red dashed circles) and produces view-consistent edits, while leaving the rest of the scene unchanged. Geometry stays sharp and textures remain faithful in unedited areas, with no noticeable artifacts.}
\vspace{-3mm}
\label{fig:more_result_1}
\end{figure}

\begin{figure}[h]
\begin{center}
\includegraphics[width=\linewidth]{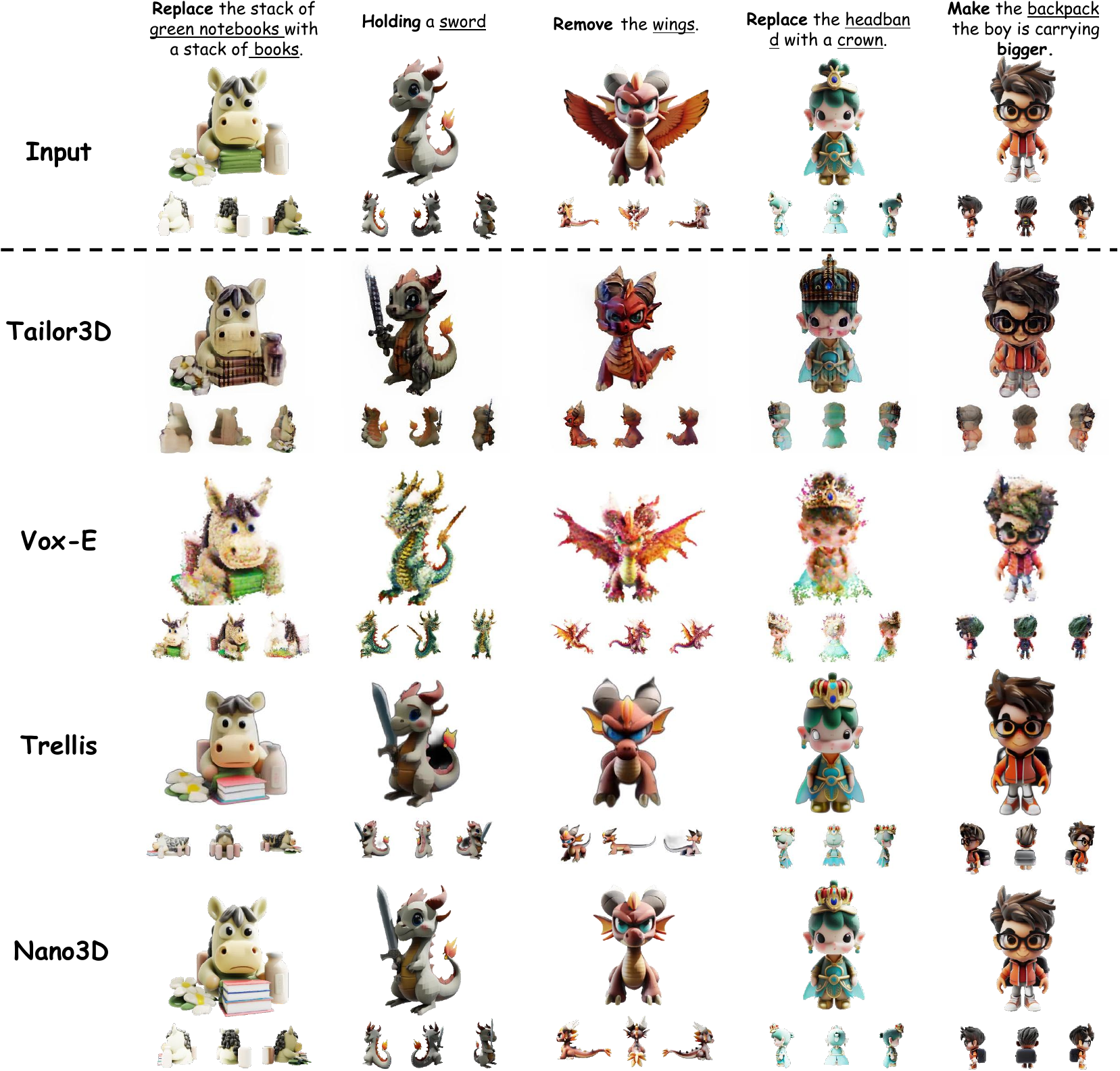}
\end{center}
\caption{\textbf{Qualitative comparison.} Our method achieves the best editing performance with faithful instruction semantic alignment and perfect original structure consistency across multi-view images.}
\label{fig:multiview}
\end{figure}

\subsection{Setup}
\textbf{Implementation Detail.} Our method is implemented on TRELLIS. The sampling step is fixed at 25, and FlowEdit is configured with $n_{max}=15$, $n_{min}=0$, and $n_{avg}=5$. The CFG guidance scales for $v_t^{\theta}(p_t)$ and $v_t^{\theta}(q_t)$ are set to 1.5 and 5.5, respectively, with $\lambda$ set to 0.5. For both Voxel-Merge and Slat-Merge, $\tau$ is set to 100. For the construction of Nano3D-Edit-100k, we employ 32 A800 GPUs for inference, utilizing the Qwen2.5-vl-72B API to generate editing instructions and Flux-Kontext to perform image editing operations. The creation of each editing pair required approximately five minutes, and empirical observations revealed two key findings: first, the vast majority of failed cases originated from errors in the image editing stage, whereas successful adherence to instructions at this stage led to a very high success rate in the subsequent Nano3D editing process; second, the predominant computational cost arose from the Flexicube module, which consumed nearly 4.5 minutes per pair, while the preceding steps required only about 30 seconds. Based on these observations and in order to further reduce computational overhead, we adopted a storage strategy in which only the SLAT (Structured Latent) representation and the voxel sum of each asset are preserved, thereby allowing users to flexibly decide whether to directly train  on SLAT or to employ Flexicube to convert SLAT into explicit GLB meshes for downstream applications. To improve dataset quality, we use Qwen2.5-VL-7B to automatically filter edited images based on instruction compliance. Non-compliant samples are returned to the pool for re-sampling.

\textbf{Baseline.} We select three representative state-of-the-art methods as baselines: Vox-E based on SDS, Tailor3D based on "multi-view editing then reconstruction", and TRELLIS, which leverages a RePaint-based method. For all baselines, we strictly follow their original implementations and use the official codebases to obtain the results reported in this paper. 

\textbf{Dataset.} Our \textbf{Nano3D-Edit-100k} dataset comprises two sources of image data: images collected from the internet and rendered views from the Trellis-500K dataset. During dataset construction, we follow the methodology of 3D-Alpaca~\cite{ye2025shapellmomni}, employing Qwen2.5-VL to automatically annotate 3D assets and classify them accordingly. We then perform class balancing across ten distinct categories, ultimately selecting 100K image samples. We select 100 representative cases from the Nano3D-Edit-100k dataset for the experiments and demonstrations in this section.

\textbf{Metric.} We systematically evaluate the edited 3D objects from three perspectives: source structure preservation, target semantic alignment, and generation quality. For source structure preservation, we assess non-edited regions against the original 3D object using Chamfer Distance (CD)~\cite{fan2017cd}. For target semantic alignment, we employ the DINO-I~\cite{caron2021dinoi} metric to quantify adherence to the target edited image. For generation quality, we use FID~\cite{heusel2017fid} on rendered multi-view images to measure fidelity and diversity.

\subsection{Main Result}
\paragraph{Qualitative Comparison.} As shown in Fig.~\ref{fig:multiview}, Nano3D not only strictly follows editing instructions but also maintains perfect structure consistency with the source 3D object across multi-view images. In contrast, Tailor3D introduces noticeable geometry distortions and appearance artifacts. Vox-E produces results that are overly blurry, smoothed, and misaligned with the target semantic. TRELLIS, though showing relative improvements, still suffers from several issues, such as local detail corruption, shape enlargement, and incorrect orientation. These findings demonstrate that our method delivers impressive and steady visual effects beyond the reach of existing methods.

\begin{table}[ht]
\centering
\begin{minipage}{0.45\linewidth}
\centering 
\caption{\textbf{Quantitative comparison.} Our method achieves the best structure consistency, semantic alignment with the target edited image, and generation fidelity.} 
\label{tab:multiview}
\renewcommand{\arraystretch}{1.1}
\begin{tabular}{lccc} 
\toprule
\textbf{Method} & CD$\downarrow$ & DINO-I$\uparrow$ & FID$\downarrow$ \\ 
\midrule
Tailor3D & 0.037 & 0.759 & 140.93 \\
Vox-E    & / & 0.782 & 117.12 \\ 
TRELLIS  & 0.019 & 0.901 & 49.57 \\
Nano3D   & \textbf{0.013} & \textbf{0.950} & \textbf{27.85} \\
\bottomrule
\end{tabular} 
\end{minipage}%
\hfill
\begin{minipage}{0.45\linewidth}
\centering
\caption{\textbf{User study.} Given that most users favored TRELLIS and Nano3D, the results for Tailor3D and Vox-E are omitted from the table for clarity. As shown in the table, our method is strongly preferred by participants, significantly outperforming TRELLIS.}
\label{tab:user-study}
\begin{tabular}{lccc}
\hline
 \textbf{Method} & \makecell{Prompt \\ Algn.} & \makecell{Visual \\ Quality} & \makecell{Shape \\ Preserv.} \\
\hline
TRELLIS  & 32\% & 21\% & 5\% \\
Nano3D     & 68\% & 79\% & 95\% \\
\hline
\end{tabular}
\end{minipage}
\end{table}

\paragraph{Quantitative Comparison.} As shown in Tab.~\ref{tab:multiview}, Nano3D outperforms all baselines, achieving the lowest CD and FID and the highest DINO-I score, indicating superior structural consistency, perceptual quality, and semantic alignment, as seen in Fig.~\ref{fig:multiview}.

\begin{figure}[h]
\begin{center}
\includegraphics[width=0.9\linewidth]{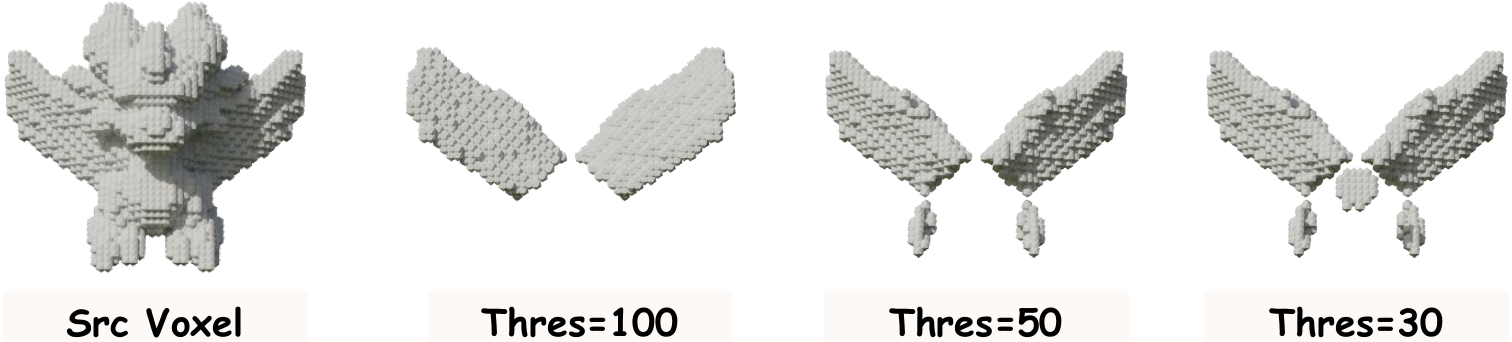}
\end{center}
\caption{\textbf{Ablation study on $\tau$.} The leftmost voxel represents the pre-editing state, with the editing instruction being to remove the wings. The three voxels on the right correspond to the masks generated during the voxel-merge stage for $\tau=100$, $\tau=50$, and $\tau=30$ (from left to right). As observed, when $\tau=100$, the detected mask most accurately aligns with the editing regions, while lower values include irrelevant non-editing areas.}
\label{fig:merge_threshold_ablation}
\end{figure}

\paragraph{User Study.} To assess editing quality and usability, we conducted a user study with 50 participants. Each round presented the original 3D object, task instructions, and results from Tailor3D, Vox-E, TRELLIS, and Nano3D. Participants selected the best method based on Prompt Alignment, Visual Quality, and Shape Preservation. As shown in Tab.~\ref{tab:user-study}, Nano3D received the highest preference across all criteria, demonstrating superior semantic alignment, visual quality, and shape fidelity. For clarity, Tailor3D and Vox-E results are omitted, as user choices mainly favored TRELLIS and Nano3D.

\paragraph{Nano3D-Edit-100k v.s. 3D-Alpaca.} High-quality 3D editing requires consistency in both 2D image appearance and 3D structure—that is, the rendered images before and after editing should remain coherent, and the 3D assets themselves should preserve structural integrity throughout the editing process. The 3D-Alpaca dataset lacks both aspects, leading to significantly lower data quality compared to ours. To quantify this, we randomly sample 500 edited pairs from each dataset and evaluate text–image alignment using CLIPScore~\cite{hessel2021clipscore} and ViLT R-Precision~\cite{kim2021vilt}. Specifically, we use a VLM to infer the caption of the edited asset based on the original asset’s caption and the editing instruction. As shown in Table~\ref{tab:nano-vs-alpaca}, our \textbf{Nano3D-Edit-100k} consistently outperforms 3D-Alpaca across all metrics.
\begin{table}[ht]
\centering
\caption{Semantic alignment comparison between \textsc{Nano3D-Edit-100k} and 3D-Alpaca.}
\label{tab:nano-vs-alpaca}
\begin{tabular}{lccc}
\toprule
 & \textbf{CLIPScore} & \textbf{ViLT R-Precision R@5} & \textbf{ViLT R-Precision R@10} \\
\midrule
\textbf{3D-Alpaca}   & 28.42 & 33.6 & 40.2 \\
\textbf{Nano3D-Edit-100k} & \textbf{39.71} & \textbf{45.3} & \textbf{52.4} \\
\bottomrule
\end{tabular}
\end{table}

\subsection{Ablation Study}
\begin{figure}[h]
\begin{center}
\includegraphics[width=\linewidth]{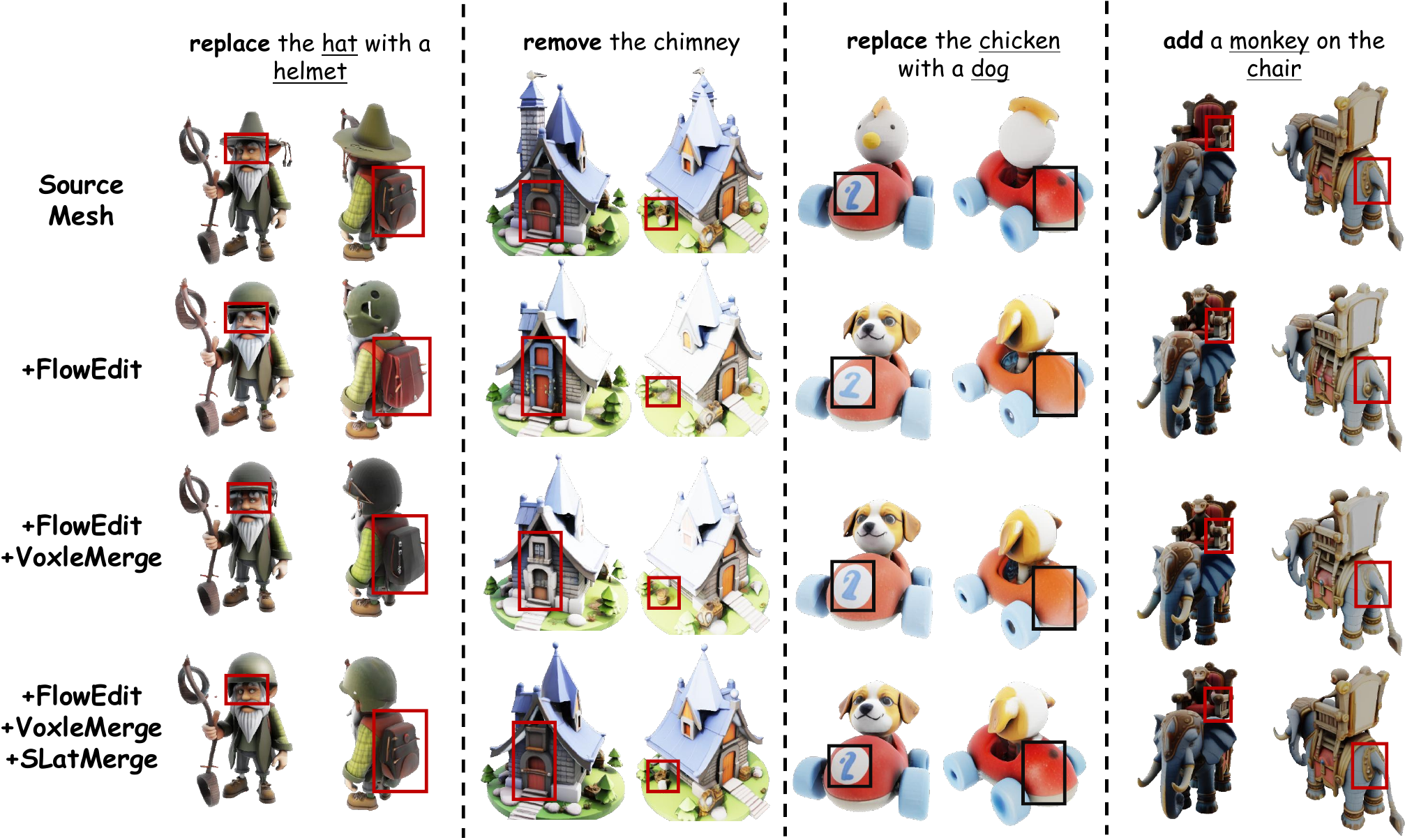}
\end{center}
\caption{\textbf{Ablation study on Voxel/Slage-Merge.} Our methods sequentially ensure geometry and appearance consistency, demonstrating their complementary roles.}
\label{fig:voxel_slat_merge_ablation}
\end{figure}

\textbf{Voxel/Slat-Merge.} We sequentially validate the effectiveness of Voxel-Merge and Slat-Merge strategies. As shown in Fig.~\ref{fig:voxel_slat_merge_ablation}, relying solely on FlowEdit leads to geometry misalignments and deformations, accompanied by missing, blurred, and distorted appearances, resulting in obvious inconsistencies with the original 3D object. Incorporating Voxel-Merge substantially improves the overall performance, restoring geometry and enhancing cross-view global consistency, but leaving appearance issues unresolved. With the additional incorporation of Slat-Merge, local visual quality is further enhanced, and appearances exhibit greater consistency before and after editing. These results indicate that our methods effectively exploit the advantage of geometry-appearance decoupling in 3D objects, ensuring more reliable consistency.
\begin{figure}[ht]
\begin{center}
\includegraphics[width=\linewidth]{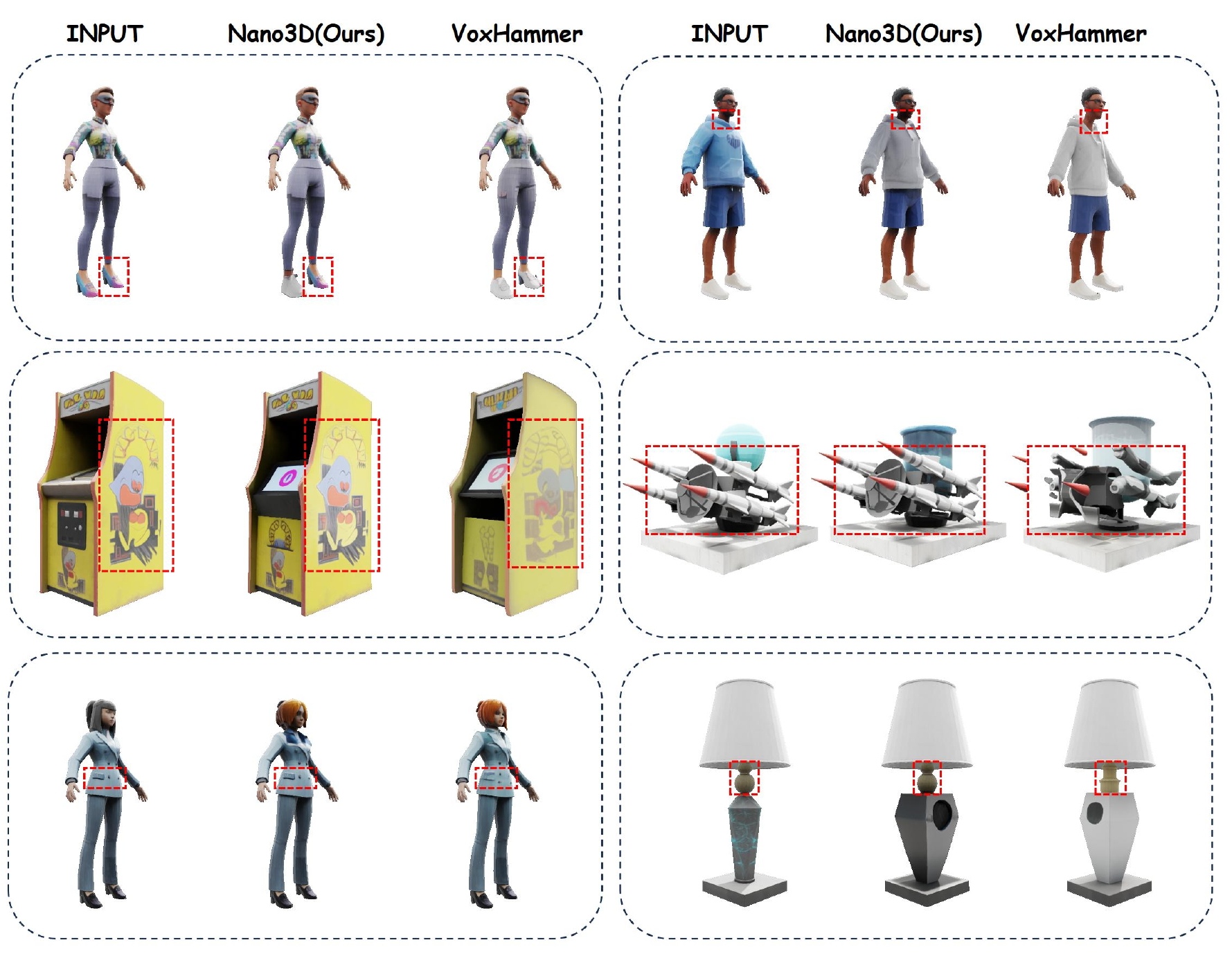}
\end{center}
\caption{\textbf{Comparison with Editing Methods Requiring Masked Input} Comparison between Nano3D and VoxHammer under identical mask-guided editing settings. Both methods receive the same bounding boxes, editing instructions, and target outputs. Nano3D preserves high consistency in non-edited regions, while VoxHammer exhibits inconsistencies compared to the input 3D assets in these areas, as evidenced by the red-bordered highlights.}
\label{fig:compare_with_vox_hammer}
\end{figure}
\textbf{Ablation on $\tau$.} We further compare different values of $\tau$, as shown in Fig.~\ref{fig:merge_threshold_ablation}. The leftmost voxel represents the pre-editing state, with the editing instruction being to remove the wings. The three voxels on the right correspond to the masks generated during the voxel-merge stage for $\tau=100$, $\tau=50$, and $\tau=30$ (from left to right). As observed, when $\tau=100$, the detected mask most accurately aligns with the editing regions, while lower values include irrelevant non-editing areas.

\subsection{Comparison with Editing Methods Requiring Masked Input}
We extend Nano3D to support user-provided masks for controlling editing regions. Experimental results demonstrate that mask-guided editing further improves Nano3D’s performance. To validate its effectiveness, we compare Nano3D with VoxHammer~\cite{li2025voxhammer}, a state-of-the-art editing method that also requires mask inputs. As shown in Figure~\ref{fig:compare_with_vox_hammer}, both methods receive identical bounding boxes, editing instructions, and target edited images. Nano3D achieves high consistency in non-edited regions, attributed to the robust capability of FlowEdit and the effectiveness of our voxel/slat merge algorithm. In contrast, VoxHammer exhibits noticeable inconsistencies in non-edited areas before and after editing. To emphasize this difference, we outline the inconsistent regions with red borders for direct visual comparison. The comparison demonstrates that our algorithm not only offers broader applicability—achieving high-consistency 3D editing without requiring user-provided masks—but also maintains state-of-the-art performance when mask inputs are provided.
\section{Conclusion}
\vspace{-1mm}
In this work, we present \textbf{Nano3D}, a training-free and user-friendly framework for localized 3D object editing, supporting operations such as object removal, addition, and replacement. By integrating FlowEdit into the TRELLIS pipeline and introducing region-aware merging strategies (Voxel/Slat-Merge), Nano3D achieves geometrically consistent and semantically faithful edits. Extensive experiments demonstrate its state-of-the-art performance across diverse editing tasks. Furthermore, we construct Nano3D-Edit-100k, the first large-scale dataset tailored for 3D editing, enabling future research on feedforward DiT-based editing models.

\textbf{Limitation.} Nano3D demonstrates strong performance in 3D editing tasks, but has the following limitations: it supports only localized edits; the VAE in TRELLIS introduces reconstruction loss; and the overall performance is constrained by TRELLIS's generative capacity. We view these limitations as important directions for future research.

\bibliography{refs}
\bibliographystyle{iclr2026_conference}
\newpage
\appendix
\section{Appendix}
\subsection{More visualization results}
\begin{figure}[ht]
\begin{center}
\includegraphics[width=\linewidth]{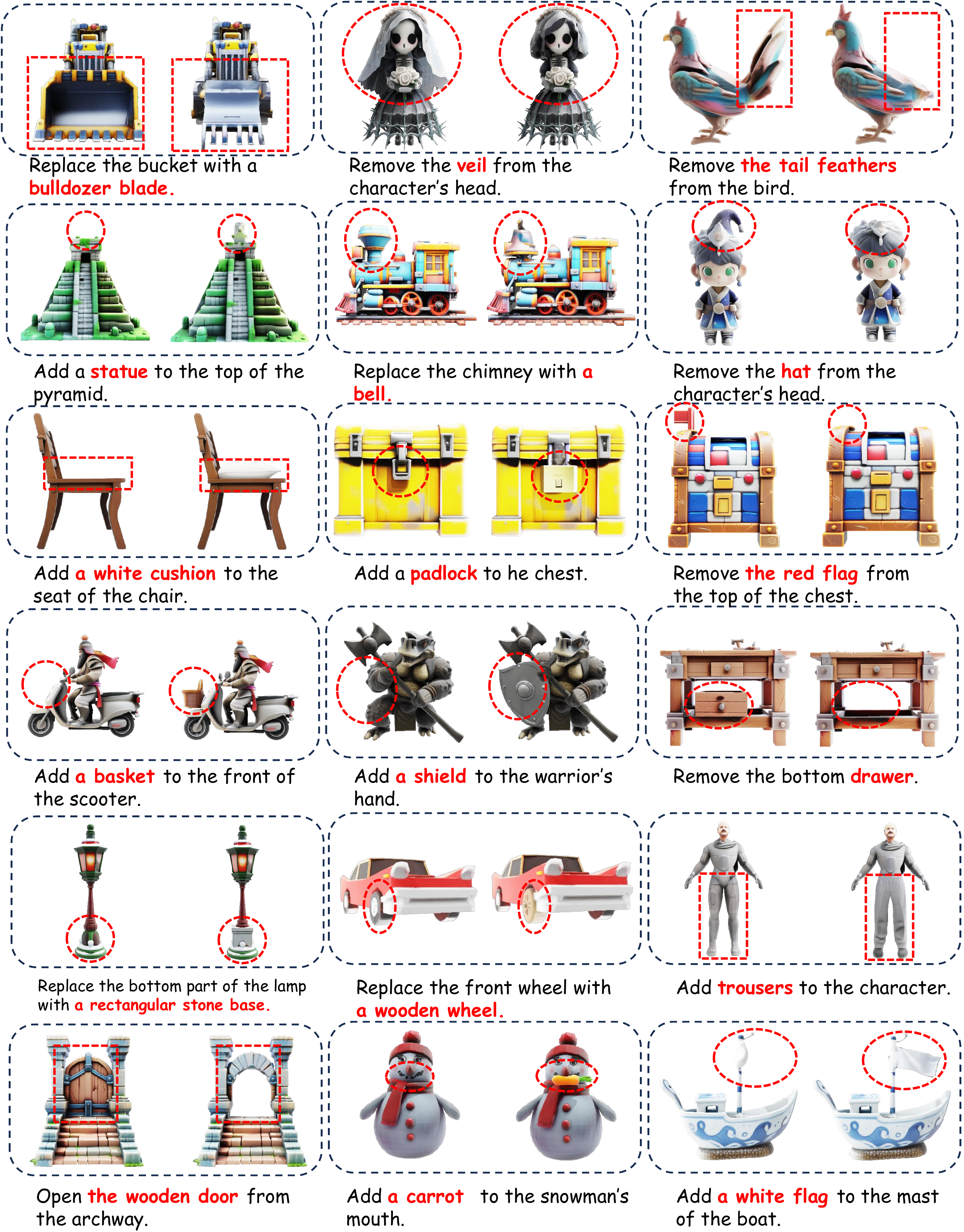}
\end{center}
\caption{We present additional editing results involving addition, removal, and replacement. Edited regions are highlighted with red dashed circles. As shown, Nano3D achieves high editing consistency, preserving geometry and texture outside the edited areas.}
\label{fig:more_result}
\end{figure}

In Fig.~\ref{fig:more_result}, we showcase additional editing results covering three types of operations: addition, removal, and replacement. As illustrated in the figure, our method, Nano3D, effectively preserves both geometric and textural consistency of the 3D assets before and after editing.
\subsection{Choice of 3D Representation: Voxel vs. VecSet}
We attempted to integrate FlowEdit into Hunyuan2.1~\cite{hunyuan3d2025hunyuan3d2.1}, but the inference results were highly unstable. When using aggressive hyperparameters, the generated 3D assets collapsed into fragmented or "mud-like" shapes. Conversely, with more conservative hyperparameters, the edited assets ignored the target condition entirely and reproduced a mesh nearly identical to the source. We believe the primary reason FlowEdit works in TRELLIS but fails in Hunyuan2.1 lies in the difference in 3D representations: TRELLIS uses a voxel-based representation, which is more local and thus compatible with localized editing methods, whereas Hunyuan2.1 adopts a vecset-based representation~\cite{zhang20233dshape2vecset}, which is more global and less suitable for transferring such localized editing techniques.
\subsection{Effect of Image Consistency in FlowEdit Editing}
\begin{figure}[!h]
\begin{center}
\includegraphics[width=\linewidth]{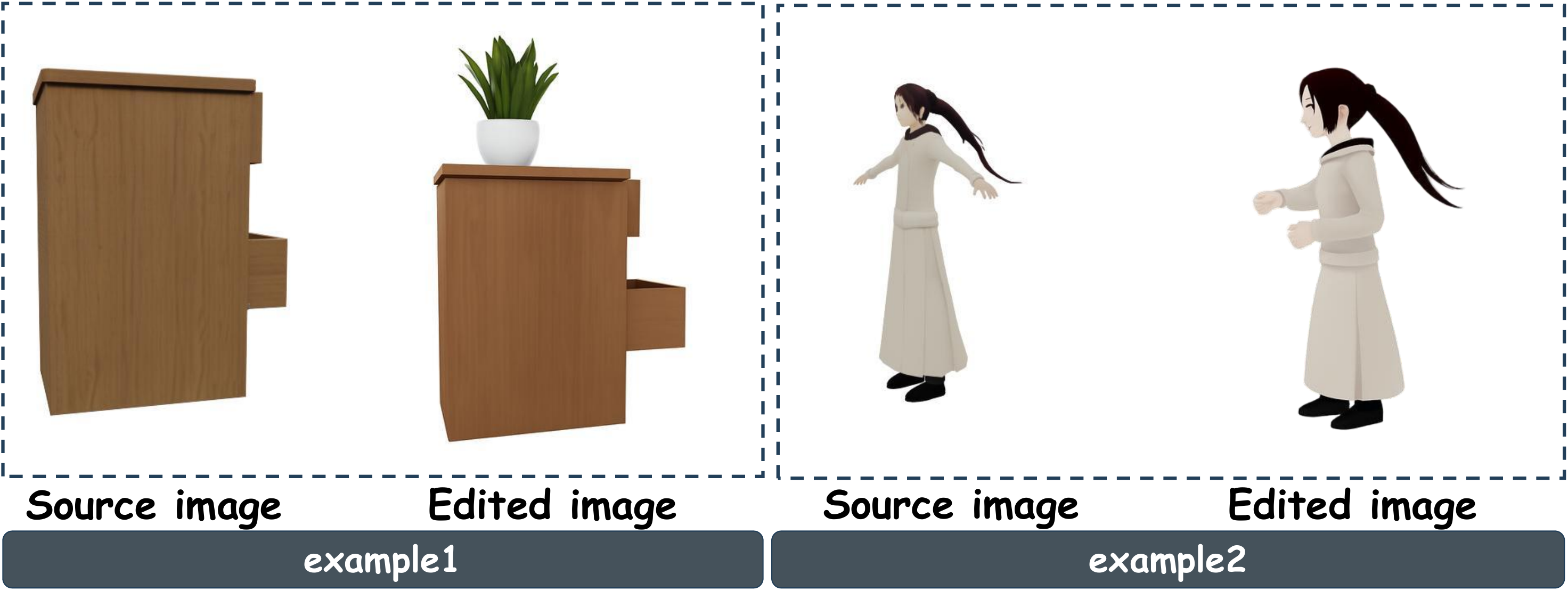}
\end{center}
\caption{A bad case sampled from the 3D-Alpaca~\cite{ye2025shapellmomni} dataset shows that its image consistency before and after editing is poorly maintained.}
\label{fig:POL}
\end{figure}

We sample several cases from the 3D-Alpaca dataset~\cite{ye2025shapellmomni}. As shown in Fig.~\ref{fig:POL}, the dataset exhibits poor 2D consistency: in the left example, the cabinet changes its position after editing, while in the right example, the character’s scale is altered. Such inconsistencies between pre- and post-edit renderings also lead to significant mismatches in the corresponding 3D assets. Following this observation, we further evaluate our Nano3D framework using the same data in Fig.~\ref{fig:POL}. The results show that under such inconsistent 2D conditions, FlowEdit fails to achieve reliable localized editing. Specifically, when n\_max is set large, the output remains nearly identical to the source asset, ignoring the target condition; when n\_max is set small, the source condition is disregarded and the results become entirely inconsistent.
\subsection{The prompt used to generate editing instruction from the rendering}
As shown in the Table.~\ref{prompt:PS}, we present an example of constructing editing instructions with a VLM. A strict template is used to constrain the VLM and prevent it from generating instructions beyond Nano3D’s capabilities.
\begin{table}[t]
  \centering
  \caption{Category distribution.}
  \label{app_tab:category_dist}
  \vspace{0.25cm}
  \begin{tabular}{r l r r}
    \toprule
    Rank & Category & Count & Share (\%) \\
    \midrule
     \rowcolor[HTML]{EFEFEF} 1 & Human & 20,755 & 20.75 \\
     2 & Weapon & 11,021 & 11.03 \\
     \rowcolor[HTML]{EFEFEF} 3 & Furniture & 10,442 & 10.45 \\
     4 & Personal Item & 10,277 & 10.28 \\
     \rowcolor[HTML]{EFEFEF} 5 & Animal & 10,186 & 10.19 \\
     6 & Vehicle & 9,376 & 9.38 \\
     \rowcolor[HTML]{EFEFEF} 7 & Building & 9,005 & 8.97 \\
     8 & Else & 6,593 &	6.60 \\
     \rowcolor[HTML]{EFEFEF} 9 & Electronic Device & 5,283 & 5.29 \\
    10 & Plant & 4,441 & 4.45 \\
    \rowcolor[HTML]{EFEFEF} 11 & electronics & 2,622 & 	5.29 \\
    \bottomrule
  \end{tabular}
\end{table}

\begin{table*}[]
\caption{The prompt used to generate editing instruction from the rendering}
\resizebox{\textwidth}{!}{%
\begin{tabular}{ll}
\hline
\textbf{Editing Action} &\textbf{Prompt} \\ \midrule
Replace & \begin{tabular}[c]{@{\ }p{0.9\textwidth}@{\ }}  Given an image, generate a short “replace” type editing instruction in the format:\\
        Replace [original object/part/pattern] with [new element]\\
    \\
    Additional Requirements:\\
        The [original object/part/pattern] must already exist in the image.\\
        It can be an entire object, a part of an object, a geometric shape, or a pattern.\\
        The [new element] should clearly differ from the original and fit naturally into the image.\\
        It can be another object, a different part, a new shape, text, or a new pattern.\\
        Avoid replacing with intangible elements (e.g., gases, smoke, light, shadow).\\
        Do not change colors — replacements must not involve altering the color of any existing element.\\
    \\
    General Rules:\\
        Keep the instruction short and clear.\\
        No extra explanation or description.\\
\end{tabular} \\ \hline
Remove & \begin{tabular}[c]{@{\ }p{0.9\textwidth}@{\ }} Given an image, generate a short 'remove; type editing instruction in the format:\\
        Remove [object/part]\\
\\
    Additional Requirements:\\
        The [object/part] must already exist in the image.\\
        It can be the whole object or a specific part of an object (e.g., handle of a cup, branch of a tree).\\
        The removal should be visually noticeable and affect the composition of the image.\\
        Avoid removing intangible elements (e.g., light, shadow, gases, smoke).\\
\\
    General Rules:\\
        Keep the instruction short and clear.\\
        No extra explanation or description.\\
\end{tabular} \\ \hline
Add & \begin{tabular}[c]{@{\ }p{0.9\textwidth}@{\ }} Given an image, generate a short “add” type editing instruction in the format:\\
        Add [element] to [location]\\
\\
    Additional Requirements:\\
        The [location] can be:\\
            an existing object in the image,\\
            a position within the image (e.g., top left, bottom center),\\
            or a specific part/position of an object (e.g., handle of a cup, roof of a house).\\
        The [element] should blend naturally into the image and not appear abrupt.\\
            It can be an object, text, pattern, or other visual addition.\\
            Avoid adding gases, smoke, or other intangible elements.\\
\\
    General Rules:\\
        Keep the instruction short and clear.\\
        No extra explanation or description.\\
\end{tabular} \\ \hline
\end{tabular}%
}
\label{prompt:PS}
\end{table*}

\end{document}